\documentclass[letterpaper, 10 pt, conference]{ieeeconf}  

\IEEEoverridecommandlockouts                              

\usepackage{graphics} 
\usepackage{amsmath} 
\usepackage{upgreek}
\usepackage{amsmath} 
\usepackage{amssymb}  
\usepackage{bm}
\usepackage{graphicx}
\usepackage{gensymb}
\usepackage{xcolor}
\usepackage{cite}
\input{Anonymize}

\title{\LARGE \bf
Selectively Actuated Fiberbots for High-precision Minimally Invasive Surgery
}

\author{Jinshi Zhao$^{1*}$, Libaihe Tian$^{1*}$, Alex Ranne$^{2*}$, Qindong Zheng$^{3}$, Yue Li$^{2}$, Xiaotong Guo$^{4}$, Ziqi Yang$^{1}$,\\ Ali Anil Demircali$^{1}$, Mohamed E. M. K. Abdelaziz$^{2}$, Bruno Gil Rosa$^{2}$,
Ferdinando Rodriguez y Baena$^{2}$,\\  and Burak Temelkuran$^{1}$
\thanks{$^{*}$ These authors contributed equally to this work.}%
\thanks{$^{1}$Jinshi Zhao, Libaihe Tian, Ziqi Yang, Ali Anil Demircali, and Burak Temelkuran are with the Department of Metabolism, Digestion and Reproduction, Imperial College London, SW7 2AZ, UK (Corresponding author: Burak Temelkuran, email: b.temelkuran@imperial.ac.uk)
}%
\thanks{$^{2}$Alex Ranne, Yue Li, Mohamed E. M. K. Abdelaziz, Bruno Gil Rosa, and Ferdinando Rodriguez y Baena are with the Hamlyn Centre for Robotic Surgery, Imperial College London, SW7 2AZ, UK}%
\thanks{$^{3}$Qindong Zheng is with the Department of Bioengineering, Imperial College London, SW7 2AZ, UK}%
\thanks{$^{4}$Xiaotong Guo is with the Department of Biomedical Engineering, University of Glasgow, G12 8QQ, UK}%
}

\begin{document}

\maketitle
\thispagestyle{empty}
\pagestyle{empty}

\begin{abstract}
Long, slender instruments are essential for minimally invasive surgery (MIS), but achieving precise distal manipulation becomes increasingly difficult when actuation extends along the entire instrument. Here, we present an electrothermally actuated fiber-based robot designed for MIS procedures, in which Joule heating is concentrated in the distal active segment. The fiber is fabricated using a thermal drawing process and uses embedded metallic wires for thermal actuation. This robotic fiber, 1.9~mm in diameter and 1.2~m in length, can precisely manipulate surgical laser fibers with positional precision below 100 \(\mu\)m. The compact size of the fiber robot allows it to pass through the working channel of a flexible endoscope, while keeping both devices mechanically uncoupled to allow independent motion. A phantom demonstration combining the robotic fiber with a da Vinci instrument conceptually demonstrated its compatibility with existing endoscopic systems for clinical applications. Our robotic fiber enables precise distal motion and holds the potential to improve clinical outcomes by enhancing precision in MIS procedures.

\end{abstract}

\section{INTRODUCTION}

\begin{figure}[!t]
    \centering
    \includegraphics[width = 0.48\textwidth]{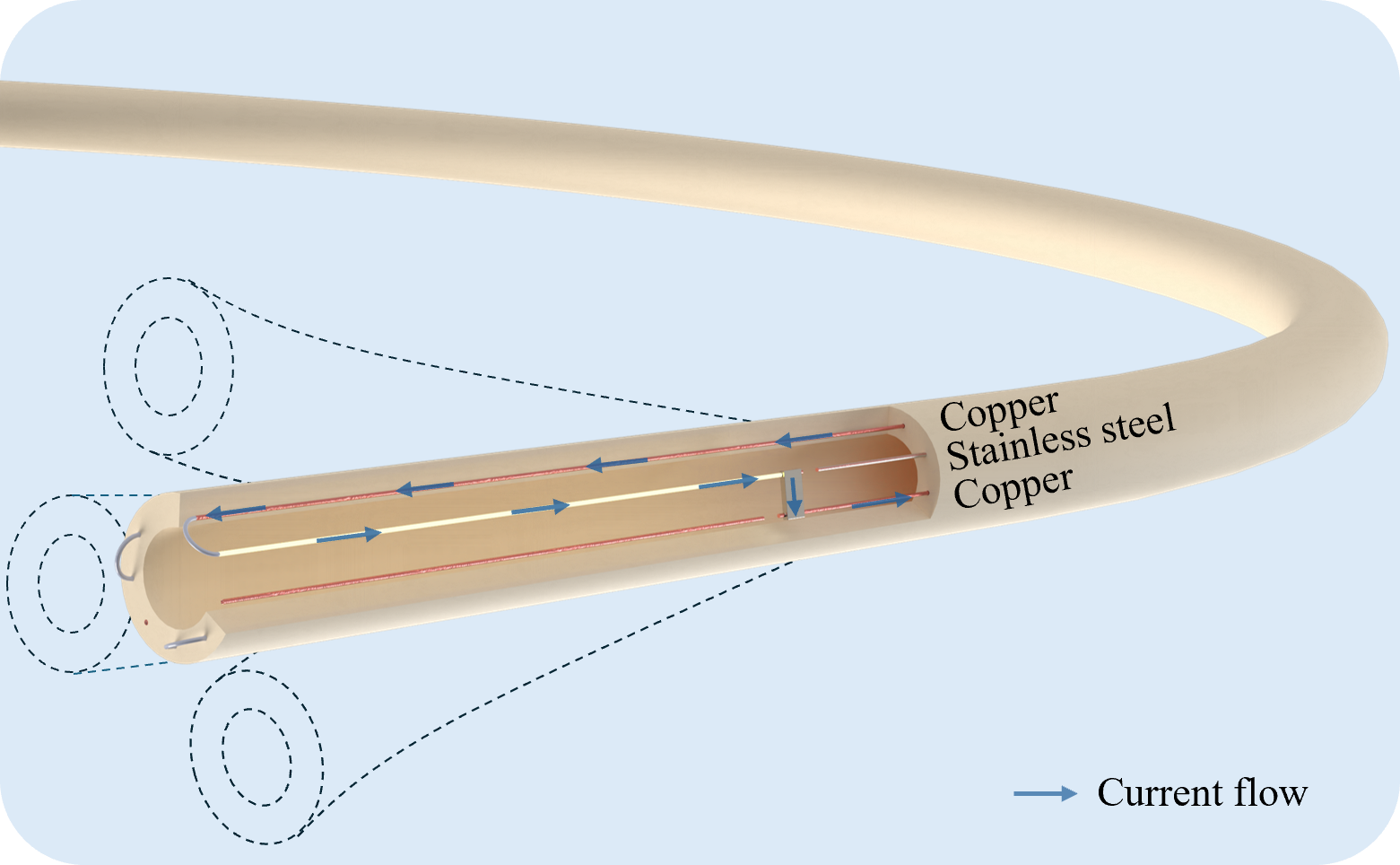}
    \caption{Schematic diagram of the selectively actuated fiberbot. }
    \label{fig: figure1}
\end{figure}

Minimally invasive surgery (MIS) has reduced the need for large surgical incisions by enabling diagnostic and therapeutic procedures through small access ports or natural anatomical pathways. Laparoscopic, endoscopic, and catheter-based procedures have consequently expanded the range of lesions that can be reached with reduced damage to surrounding tissues. Robotic technologies can further enhance these procedures by improving instrument dexterity, motion precision, and operator control within confined anatomical environments \cite{Clinicaleffectiveness, doi:10.1126/scirobotics.abi8017}.

Many minimally invasive procedures require long and slender instruments to pass through access ports, flexible endoscopes, or tortuous anatomical pathways before reaching the target site. Although this geometry facilitates access to deep anatomical regions, it makes precise distal manipulation challenging. Conventional instruments are commonly controlled from their proximal ends, and their distal response can be affected by shaft deformation, friction, backlash, and interactions with the surrounding anatomy or the working channel of an endoscope. These effects become more pronounced as the instrument length and pathway complexity increase, limiting the ability to position or scan a small distal end-effector accurately. Miniaturized continuum and soft robotic instruments offer a promising means of introducing active steering and fine manipulation at the surgical site while retaining the flexibility required for minimally invasive access \cite{9779312}.

In this paper, we present the next generation of electrothermally actuated fiber robots (Fiberbots). These robots take the form of a long (1.2~m in length) and slender (1.9~mm in diameter) catheter, with an electrothermally actuated tip capable of achieving sub-millimeter precision motion. A  selective actuation mechanism was developed by localizing the high-resistance heating element to the distal active section. The fiberbot avoids dissipating substantial electrical power along mechanically passive portions of the shaft. Fiberbots are designed to fit through the working channels of off-the-shelf endoscopes, enabling seamless integration into an operating theater. In Sect. \ref{Section: Related works}, we begin by investigating common existing designs for laparoscopic robots used in MIS surgery, while highlighting their limitations that need to be addressed. Then in Sect. \ref{Section: Methods}, we present a new fiberbot design, highlighting its novelty in selective actuation. Using thermal fiber drawing technology, a bespoke polymer multi-lumen fiber was fabricated. This technique provides the capabilities to rapidly prototype devices with complex cross-sectional designs, while mitigating the difficulties faced by conventional extrusion or molding techniques. This section is followed by a summary of circuit assembly. Finally, in Sect. \ref{Section: Experimental evaluation}, we performed motion characterization to evaluate the fiberbot’s precision, repeatability, and reachable workspace. We also conducted temperature characterization to examine the temperature profile under different current levels and validate the efficiency and clinical viability of the selective actuation mechanism. To demonstrate system-level feasibility, we integrated the fiberbot with a da Vinci surgical instrument and used it to perform a moving task in an endoluminal phantom. The proposed selective-actuation architecture addresses a fundamental scaling limitation of electrothermally actuated fiberbots \cite{abdelaziz2024fiberbots}. By localizing the high-resistance heating element to the distal active section, the fiberbot avoids dissipating substantial electrical power along mechanically passive portions of the shaft, thereby improving its suitability for high-precision minimally invasive surgery.

\section{RELATED WORKS} \label{Section: Related works}

 \subsection{Manipulators for Minimally Invasive Surgery}
Miniaturized manipulators for endoscopic and catheter-based interventions can be broadly distinguished by the location and type of their actuators. Tendon-driven systems use proximally located motors to transmit forces through cables along the instrument body. These systems have achieved substantial clinical maturity, but their distal response can be affected by tendon friction, wire slack, backlash, and hysteresis \cite{cho2024accounting, fang2023design, lee2021non, 11197914}. Concentric-tube robots generate three-dimensional motion by translating and rotating nested precurved tubes. Although they provide a compact structure, their behavior depends on tube interactions and elastic stability, and abrupt configuration changes can occur under certain designs and loading conditions \cite{harvey2020novel, rox2020mechatronic, boyer2024high, donat2021real, gilbert2015elastic}.

Magnetically actuated catheters place a magnetic component at the catheter body and generate steering forces using an externally applied magnetic field \cite{FDA_K193147, torlakcik2021magnetically}. This approach removes mechanical transmission elements from the catheter body but requires dedicated magnetic-field-generation infrastructure.
Fluidic-driven catheters achieve bending by pressurizing internal chambers, which deforms the robot’s compliant structure \cite{gopesh2021soft}. These catheters can achieve large bending angles, although fabrication and control become increasingly challenging as channel dimensions decrease \cite{NGUYEN2023114380, https://doi.org/10.1002/adrr.202500197}. Shape-memory actuators operate through a temperature-induced phase transformation: heating drives the material toward its memorized shape, while cooling allows it to recover \cite{kim2023shape}. Although they offer a high force-to-volume ratio and compact construction, their nonlinear, temperature-dependent, and hysteretic behavior can complicate accurate open-loop positioning \cite{tung2008laser, li2024design, oda2024shape}.

 \subsection{Thermally Drawn Robotic Fibers}
Thermal drawing enables the continuous fabrication of polymer fibers containing internal channels and embedded functional materials  \cite{3DPrintPreform}. While it was originally introduced in the telecommunications industry to create silica-based optical fibers, researchers have recently adapted this method for prototyping medical devices \cite{zheng2026mappingsheaththermallydrawn}. 
 
 A previous study \cite{abdelaziz2024fiberbots} developed a fiberbot for laparoscopic use by powering the entire 40~cm fiber length to generate precise motion at the distal tip, demonstrated in an in vivo porcine model. However, in that work, only the distal 12~cm contributed to motion, while the remaining length was clamped within a rigid tube to prevent unintended movement. This approach resulted in powering the entire length of the instrument. Here, we present an improved iteration of the system, consisting of a new device design that enables selective actuation by actuating a targeted length at the tip of the fiber rather than the entire device. This approach is more suitable for MIS applications that require long instruments, reducing the power required for actuation and Joule heating on the overall device.

\section{METHODS} \label{Section: Methods}
\subsection{Selectively Actuated Robotic Fiber Design}

The fiberbot body design consists of a tubular polymeric structure with an outer diameter of 1.9~mm and an inner diameter of 1~mm, adapted to accommodate commercially available fiber-based surgical instruments, such as surgical laser fibers. Nine side lumens are uniformly distributed around a central channel to accommodate metallic wires. Three triplets of  60~\(\mu\)m metallic wires, arranged as 'copper–stainless steel–copper', were embedded within these side channels for actuation and power transmission (Fig.\ref{fig: figure1}). For actuation purposes, in each wire set, the stainless steel wire and one of the copper wires were electrically connected at the fiber's distal tip. At a specified distance along the fiber, the stainless steel wire was electrically connected to the other copper wire.

The resistivity of copper ($\rho_\mathrm{c} = 1.68 \times 10^{-8}\,\Omega\cdot m$) is approximately forty times lower than that of stainless steel ($\rho_\mathrm{s} = 6.9 \times 10^{-7}\,\Omega\cdot m$). Consequently, when a voltage is applied to the proximal end of two copper wires, the electrical power mainly concentrates in the stainless-steel wires due to their high resistivity, thereby forming a selectively actuated segment.

\begin{figure}[!h]
 \hspace*{-0.39cm} 
    \centering
    \includegraphics[width = 0.51
    \textwidth]{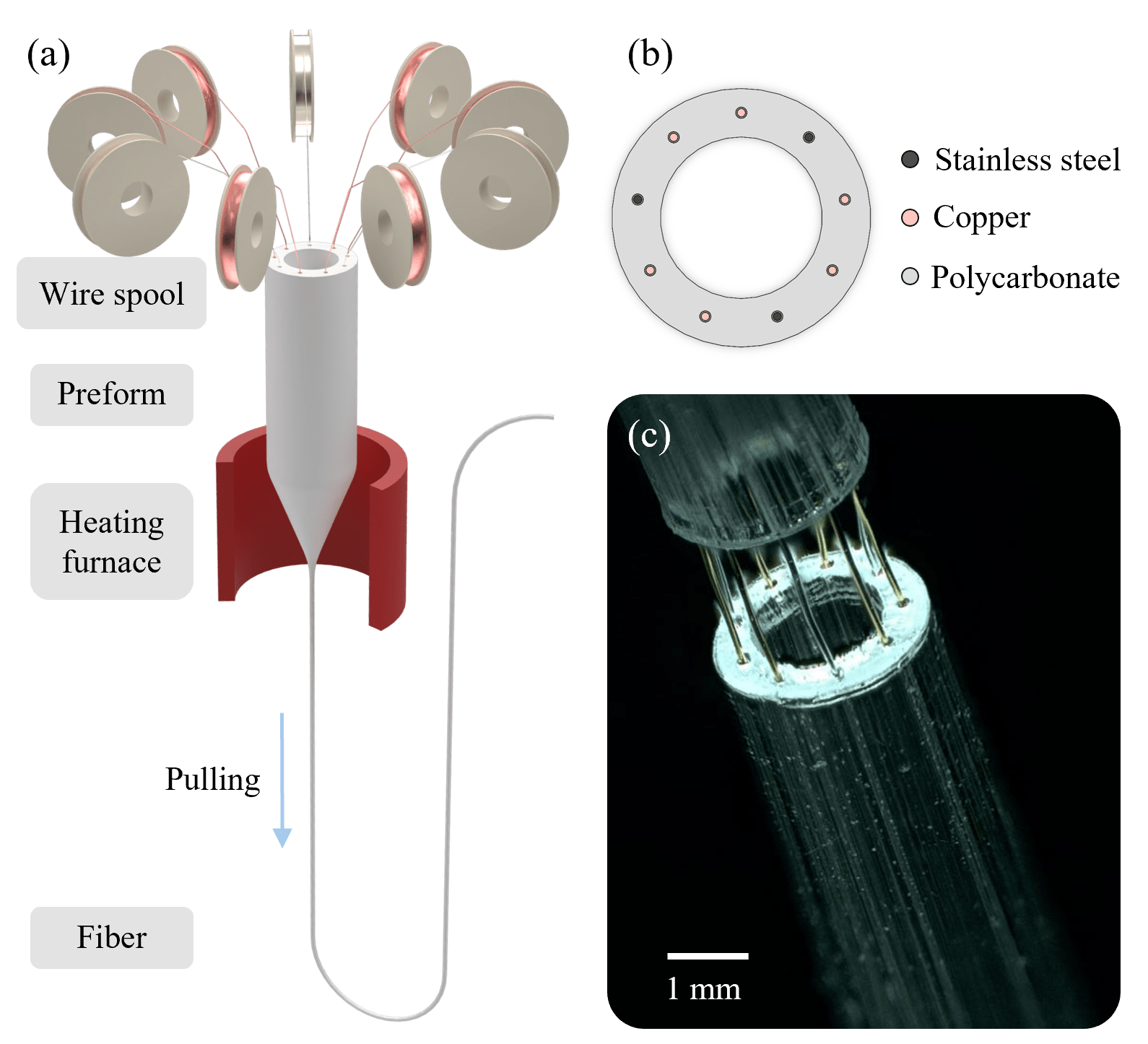}
    \caption{Fiber fabrication. (a) Schematic diagram of the thermal drawing process, showing wire spools feeding wires during drawing. (b) Cross-sectional design of the fiber. (c) Microscopic image of the fiber cross-section with exposed wires.}
    \label{fig: figure2}
\end{figure}

\subsection{Robotic Fiber Fabrication}

\subsubsection{Thermal drawing}
Thermal fiber drawing is a fabrication process to create high-aspect ratio devices. Its capabilities of retaining catheter design, while mitigating the drawbacks of extrusion (notably by lowering the cost of manufacturing and reducing tooling costs), make it a suitable alternative for this work \cite{abdelaziz2024thermally, leber2023highly,abdelaziz2024fiberbots}.

The fiber preform, a macroscopic version of the fiber design in the form of a polymer tube (15 cm in length and 4 cm in diameter) with nine side lumens uniformly distributed along its wall, was 3D-printed using polycarbonate (PC). This 3D-printed fiber preform was placed in a vacuum oven for two days to remove residual moisture. The preform was then fixed onto a preform-holding rod made from polyether ether ketone (PEEK). Additionally, a platform containing nine metallic wire spools was mounted onto the preform-holding rod above the preform. The wires from the wire-feeding platform were manually threaded through their corresponding lumens from top to bottom and then tied together in a knot. This method ensures the wires remain properly aligned and embedded within the fiber during the thermal drawing process. Subsequently, the preform assembly, including the holding rod and wire-feeding platform, was suspended from a downward-feed stage on the thermal draw tower, positioned above a three-zone heating furnace, with temperatures of 120~°C/190~°C/85~°C for the top, middle, and bottom zones, respectively.

During the thermal drawing process, the preform is heated above the glass transition temperature (in the case of PC, this is around 112~°C), which allows its body to soften but remain viscous. Weights were hung on the distal end of the preform, such that when the preform reaches the glassy state, it allows bite-off (narrowing of the preform to its final required dimension) to take place. Then, the distal end of the preform is discarded, and the fiber is fed into a rotating capstan such that the fiber is pulled continuously with a constant tension (Fig.\ref{fig: figure2}(a)).

\subsubsection{Circuit assembly}
Through thermal drawing, we created multi-lumen, high-aspect-ratio tubing with many embedded wires. 
In our fiber, we placed two copper wires beside each steel wire, then distributed three of these wire triplets evenly around the catheter circumference. To actuate the steel wire, it needs to be connected both on the proximal end and at the distal end. Practically speaking, only a small segment of the stainless steel wires, 10 cm long, needs to be actuated. To facilitate this, the proximal end connection is established by exposing several small windows of size 3~mm by 1~mm on the fiber surface using femtosecond laser micro-machining (Lasea Belgium, Seraing, Belgium), exposing a copper-steel wire pair (one of the copper wire and steel wire), which can be connected. Precise manipulation of the fiber is enabled via an automated x-y stage (with a resolution of 0.001~mm), as well as a custom-built positioning system that rotates the catheter about its central axis (with a resolution of 0.0879°)

On the distal end, the other copper wire and the steel wire are exposed by gently pulling them out of the fiber body, then first twisted together, before the wires are rigidly bonded together using laser spot-welding at 120 V (Coherent, Saxonburg, Pennsylvania). Once welded, the wires are protected using a thin layer of Kapton tape.

\begin{figure}[!h]
    \centering
    \includegraphics[width = 0.48\textwidth]{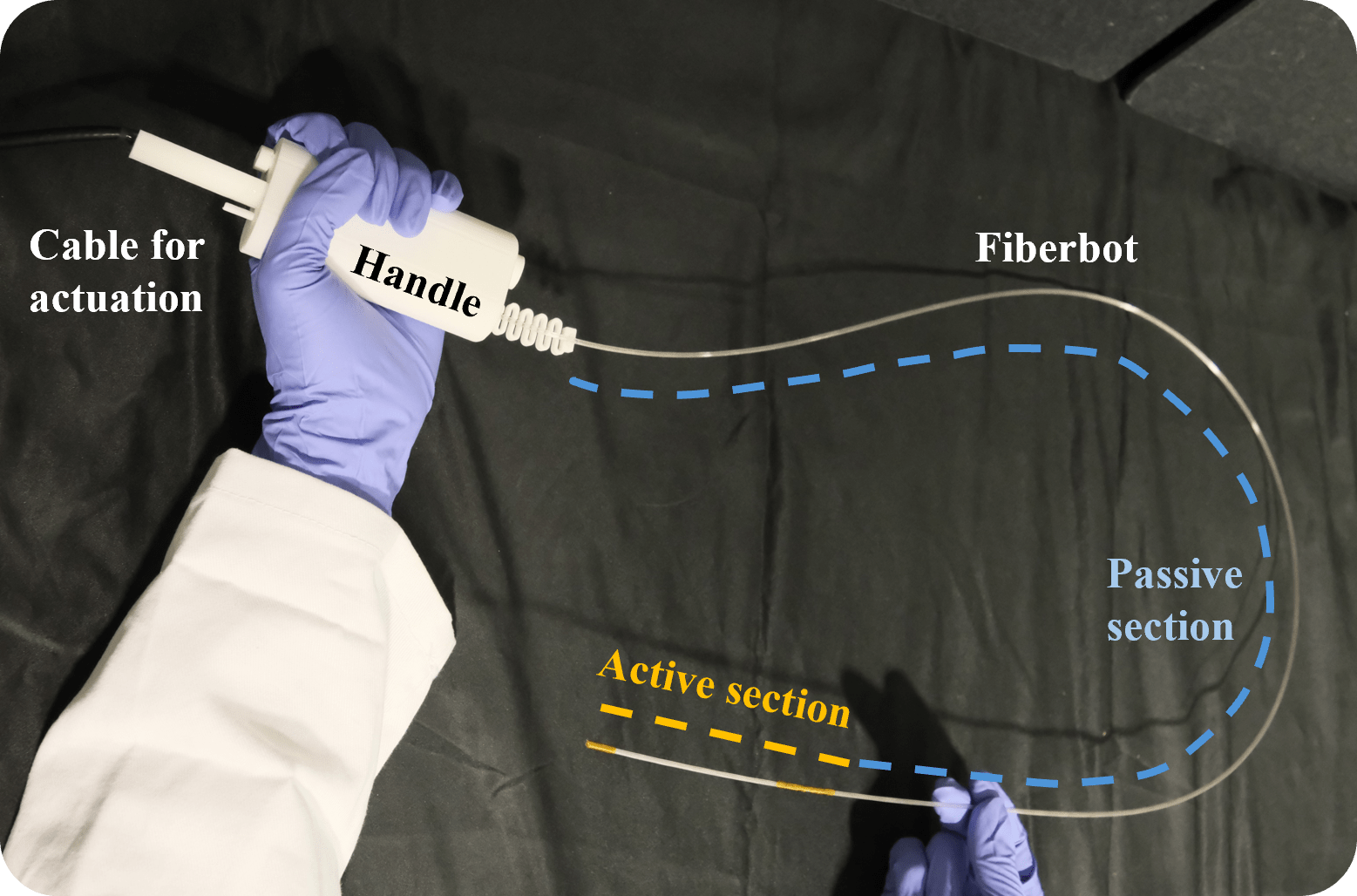}
    \caption{Selectively actuated Fiberbot with handle. }
    \label{fig: figure3}
\end{figure}

A handle was designed to securely fix the fiberbot's position and ensure a stable connection for the actuation cables. The handle features an elliptical cross-section (50~mm major axis × 30~mm minor axis) and a length of 10 cm, providing surgeons with a comfortable grip during operation. A spring-shaped structure at the distal end of the handle prevents fiber breakage caused by sharp bending at the handle edge. The proximal end of the handle includes the actuation power cable plug and the exposed proximal end of the fiberbot, allowing fiber-based surgical devices to be inserted through the central working channel. The handle was 3D-printed using polylactic acid (PLA) (Fig.\ref{fig: figure3}).

\section{EXPERIMENTAL EVALUATION} \label{Section: Experimental evaluation}

\subsection{Actuation setup}
To control and actuate the fiberbot, the adapter connected to the fiber was interfaced with a buffer circuit consisting of six operational amplifiers (op-amps). These op-amps provided voltage amplification and impedance matching between the control system and each of the actuation wires. The input of the buffer circuit was connected via jumper wires to the digital-to-analog converter (DAC) ports of a real-time controller (cRIO-9025, National Instruments, USA). The control system was developed within the LabVIEW environment (Fig.\ref{fig: figure5}(a)) \cite{abdelaziz2024fiberbots}.

\subsection{Motion Characterization}
The characterization was performed on a 10 cm-long fiberbot with only an active section. To monitor the tip position of the fiberbot, a single-mode optical fiber was inserted through the central lumen of the fiberbot. A red LED was coupled to the proximal end of this optical fiber, allowing illumination to be observed at the fiber tip. The position of the fiberbot tip was monitored using a digital microscope (VHX-6000, Keyence, Japan) placed in front of the fiber, tracking the red dot emitted from the optical fiber \cite{abdelaziz2024fiberbots}.

 To verify the system's linearity within the 2D workspace, we applied 500 sets of random input voltages to the actuation wires. For each voltage set, two of the three wire pairs were randomly selected and assigned different random voltages, while the remaining pair was set to zero. Because displacement is approximately proportional to electrical power and therefore to $V^2$ for a constant resistance, random samples were generated uniformly in the power domain and converted to the corresponding voltages using $V=\sqrt{PR}$. This produced a more uniform distribution of target positions within the workspace. We applied these voltage sets sequentially to the fiberbot, holding each set for 20 seconds to allow the fiber to reach a steady state. Between consecutive voltage sets, a waiting period of 40 seconds was implemented, during which all voltages were set to zero.

\begin{figure}[!h]
    \centering
     \hspace*{-0.35cm} 
    \includegraphics[width = 0.5
    \textwidth]{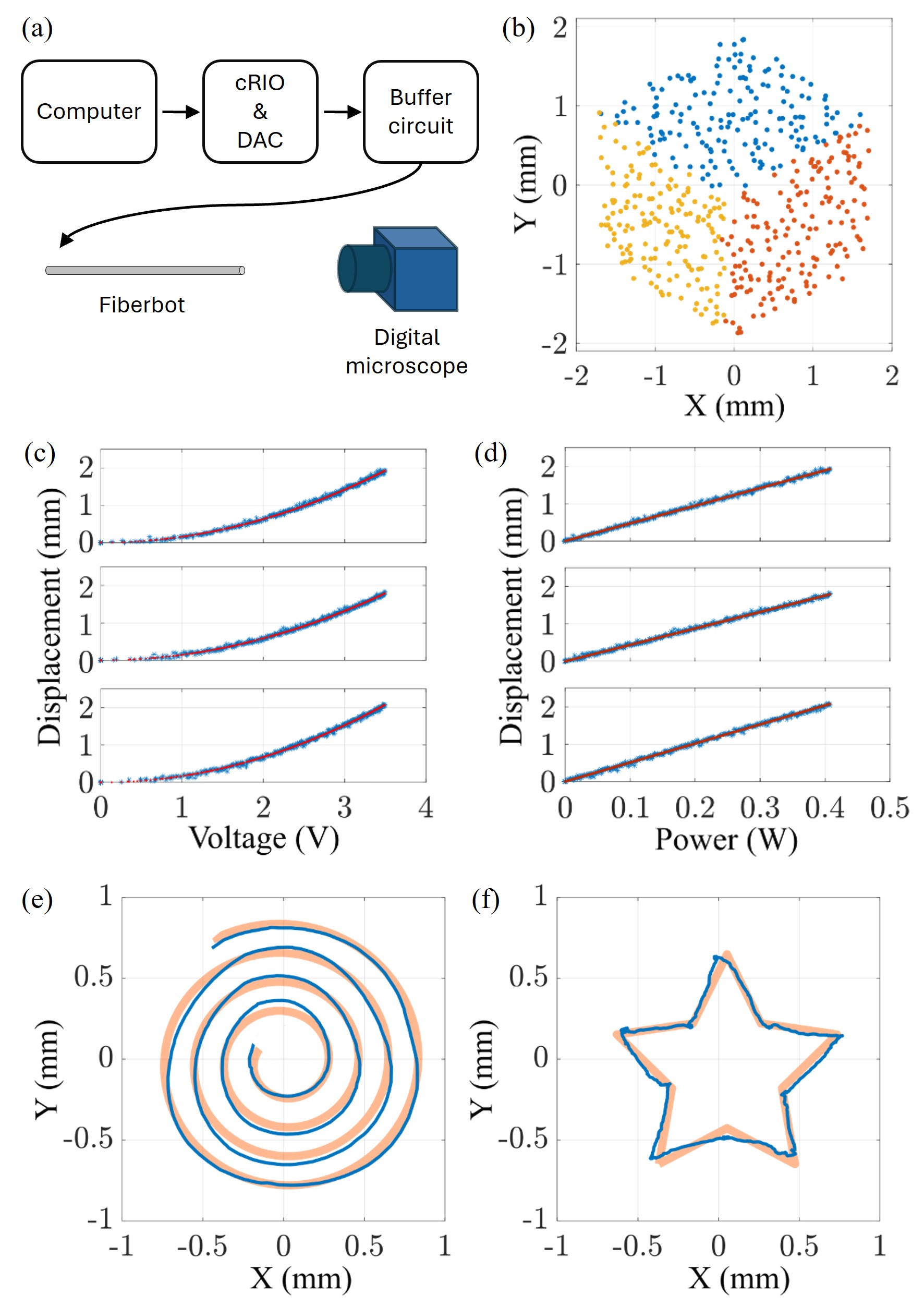}
    \caption{Fiberbot motion characterization. (a) Fiberbot actuation and characterization setup. (b) Fiberbot tip positions resulting from 500 random input voltage sets. Each color represents one of the three possible combinations of two simultaneously actuated wire sets. (c,d) Relationship between applied voltages (c), powers (d) and the vector magnitude of the fiber tip position along their corresponding actuation directions. (e,f) Measured (blue) versus desired (orange) spiral and star shape paths. }
    \label{fig: figure5}
\end{figure}

The vector from the fiber’s origin to each recorded fiberbot position was calculated and decomposed along the two actuation directions corresponding to the actuated wire pairs. We expressed each displacement vector as a linear combination of the two actuation-direction vectors, yielding the displacement components along each direction. The relationships between applied voltage (Fig.\ref{fig: figure5}(c)) and power (Fig.\ref{fig: figure5}(d)) and the corresponding displacement components were plotted, validating the linear relationship between input power and fiber displacement.

\subsection{Path Following Control}
The path following control was achieved by inputting a sequence of 2D coordinates on a specific path that was generated using MATLAB. For each coordinate, we calculated the vector from the origin to the target position and determined its component along each corresponding actuation direction. We then calculated the required power for each wire using the linear relationship derived from characterization results. The required voltages for each position were calculated based on these power components and the wires' resistance.
Each calculated voltage was then applied to the corresponding wires for 100 ms. The fiber motion was recorded using the digital microscope setup described previously. Two paths—a star shape and a spiral shape—were tested (Fig.\ref{fig: figure5}(e) and (f)). The maximum positioning error for these two paths was below 100 \(\mu\)m, confirming the precision of the fiberbot's motion under the open-loop control algorithm.

\subsection{Temperature Characterization}
To further understand the temperature distribution across the fiberbot’s cross-section, we positioned a thermal camera (FLIR A400, Teledyne FLIR LLC, USA) equipped with a two-times magnification close-up lens (T300238) in front of the fiber to observe temperature changes during actuation.

To investigate the relationship between applied electrical power, temperature, and tip displacement, a power with a 0.1 W step size was incrementally applied to a single copper-stainless steel wire pair. At each step, the voltage was applied for 20 seconds, allowing sufficient time for thermal expansion and fiber deformation to stabilize. Fig.5(a) shows thermal camera images at each voltage step. The fiber's displacement, maximum and minimum temperatures, and temperature differences across the fiber's cross-section were recorded (Fig.5(b) and (c)). The tip displacement and temperature rise increased approximately proportionally to the square of the applied voltage.

To validate the selective actuation mechanism, we captured a side view of a selectively actuated fiber, including both active and passive sections. The thermal camera lens was changed to one with a longer focal distance and larger field of view (T300240, Teledyne FLIR LLC, USA) (Fig.5(d)). The results (Fig.5(e)) indicate that while the active section of the fiber experienced a temperature increase of approximately $28.5\,^{\circ}\mathrm{C}$, the passive section increased by only around $1.4\,^{\circ}\mathrm{C}$, confirming that heat was predominantly generated in the active section of the fiber.

\begin{figure}[!h]
    \centering
     \hspace*{-0.25cm} 
    \includegraphics[width = 0.48
    \textwidth]{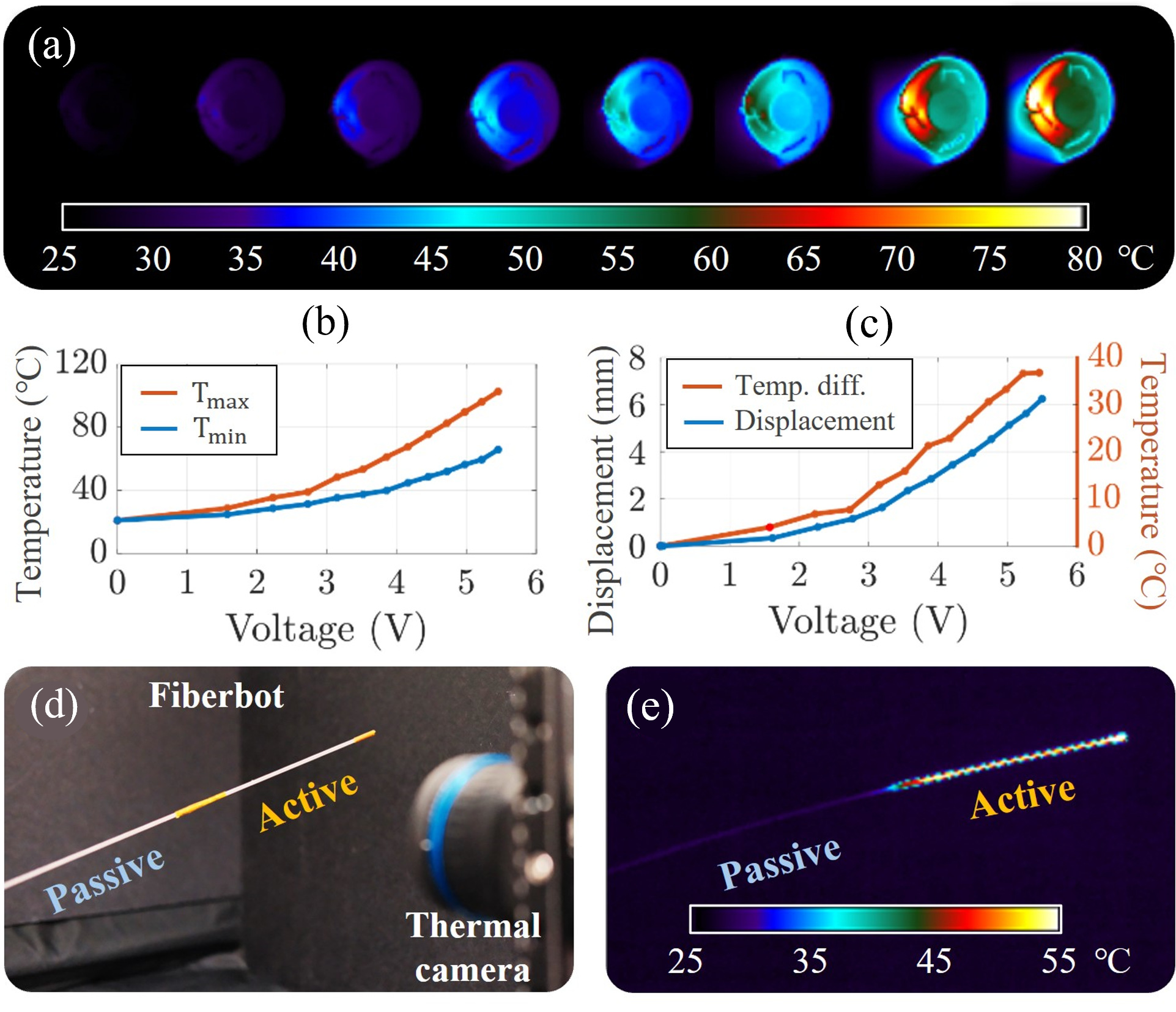}
    \caption{Temperature characterization. (a) A series of thermal imaging camera images showing the temperature distribution of the fiber at various applied voltages. (b) Maximum (orange) and minimum (blue) temperatures at various applied voltages measured using a thermal camera. (c) Fiber tip displacement (blue) and temperature difference (orange) between maximum and minimum temperatures under various applied voltages. (d,e) Thermal camera setup and corresponding image showing a side view of the selectively actuated fiber.}
    \label{fig: figure5temp}
\end{figure}

\subsection{Actuation Power Analysis}
To quantify the power-saving benefit of selective actuation, we compared the proposed design with the actuation architecture used in the previous work \cite{abdelaziz2024fiberbots}. For a fair comparison, the distal active-section length was fixed at 0.10 m, while the length of the proximal passive section was varied from 0 to 1.0 m. In both cases, the distal 0.10-m section was assumed to generate the Joule heating required to produce a 2-mm tip displacement. The corresponding actuation current was obtained from the experimentally characterized relationship between electrical power and tip displacement shown in Fig. 4(d).

The total resistance was calculated by summing the resistances of all wire segments along the current path. Contact resistance and the temperature dependence of electrical resistivity were neglected. In the selectively actuated fiberbot, the high-resistance stainless-steel wire is restricted to the distal active section, whereas current is transmitted through the passive section primarily by low-resistance copper wires. In contrast, in the architecture used in the previous work \cite{abdelaziz2024fiberbots}, the high-resistance actuation wires extend along the entire fiber. Although the proximal section is mechanically constrained by a rigid supporting structure and therefore does not contribute to tip motion, it still carries current and dissipates electrical power as Joule heat.

As shown in Fig. 6(c), the required power of the selectively actuated fiberbot increases only moderately, from approximately 0.40 to 0.75 W, as the passive-section length increases from 0 to 1.0 m. By comparison, the required power for the previous architecture increases substantially, reaching approximately 6.6 W for a 1.0 m passive section. At this length, the proposed selective-actuation design requires only approximately 11\% of the power required by the previous architecture, corresponding to an approximately 89\% reduction in total electrical power. These results demonstrate that mechanically constraining an electrically heated section prevents its motion but does not eliminate its power consumption. Selectively confining the high-resistance heating element to the distal active section therefore substantially improves the electrical efficiency of long fiberbots, with the benefit becoming increasingly pronounced as the passive-section length increases.

\begin{figure}[!h]
    \centering
     \hspace*{-0.25cm} 
    \includegraphics[width = 0.48
    \textwidth]{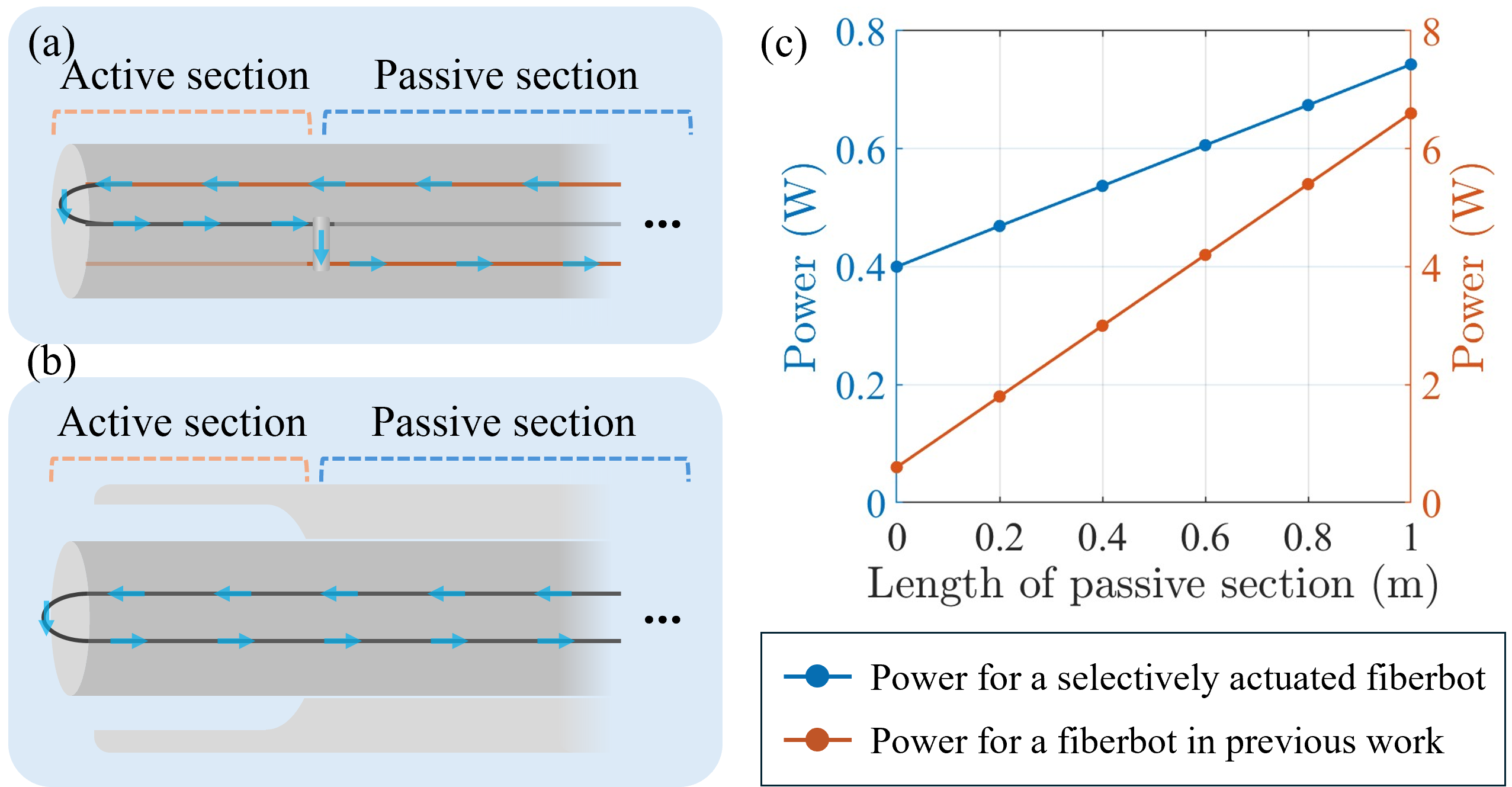}
    \caption{Comparison of the required actuation power between the selectively actuated fiberbot developed in this work and the fiberbot reported in the previous work \cite{abdelaziz2024fiberbots}. (a) Schematic of the working principle of the selectively actuated fiberbot developed in this work. (b) Schematic of the actuation approach used in the previous work. (c) Calculated total actuation power required to achieve a 2 mm tip displacement for different passive-section lengths.}
    \label{fig: figure6}
\end{figure}

\subsection{Endoluminal Phantom Demonstration}

\begin{figure}[!h]
    \centering
    \vspace*{0.3cm} 
     \hspace*{-0.3cm} 
     
    \includegraphics[width = 0.48
\textwidth]{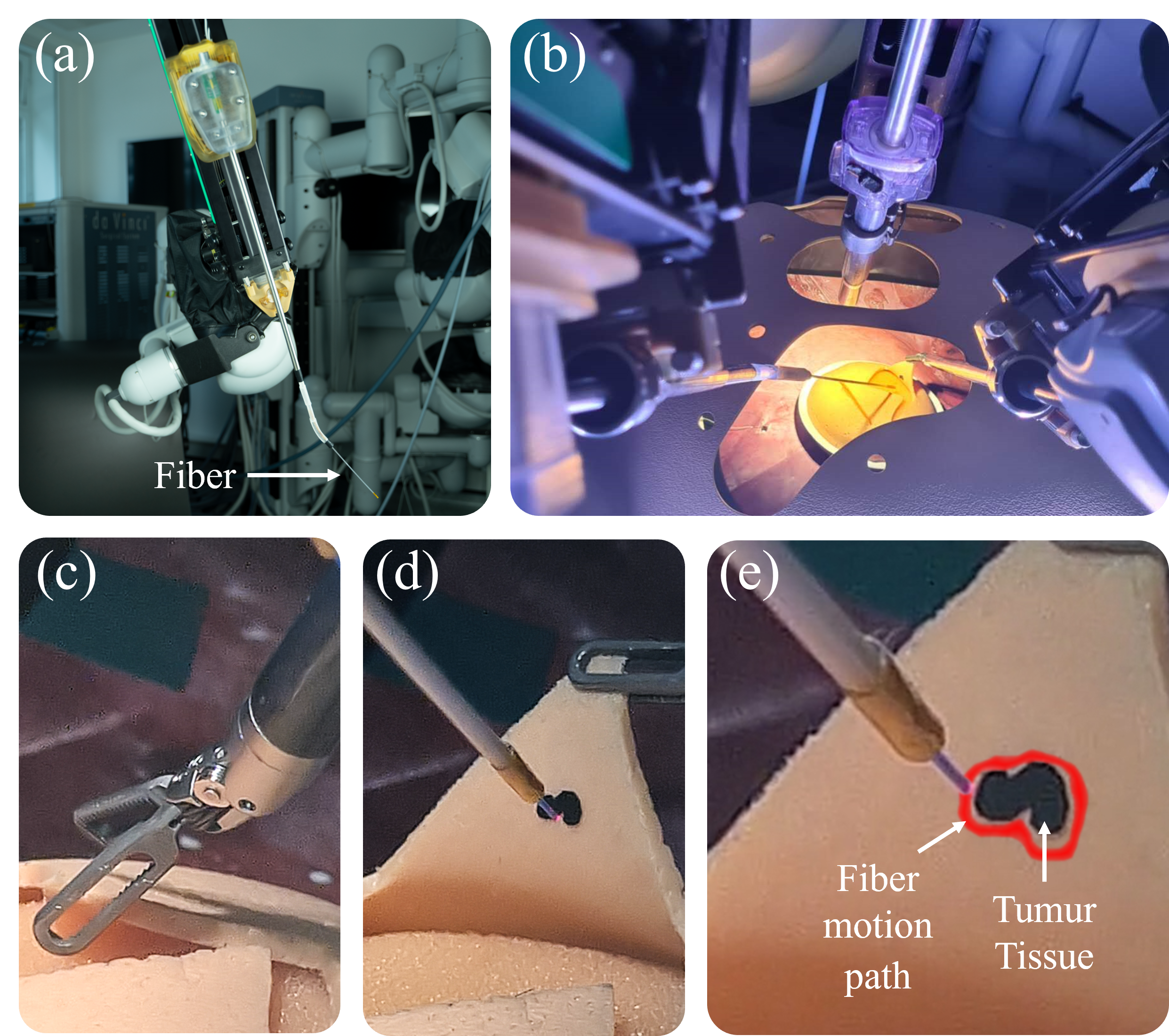}
    \caption{The da Vinci® surgical instrument integration and endoluminal phantom demonstration. (a) 
The integration between the da Vinci® surgical robot and the 
fiber actuator. (b) Experimental setup of the abdominal phantom 
study. (c) Retraction of the phantom surface layer using a grasper to expose the simulated tumor. (d) Fiberbot tip scanning along the simulated tumor boundary, with an optical fiber inserted through its central lumen. (e) Fiberbot tip motion path around the simulated tumor, indicated by the red trace.}
    \label{fig: figure7}
\end{figure}

To evaluate teleoperation in a more clinically relevant setting, we integrated the partially actuated robotic fiber into a laparoscopic device adapted for use with the da Vinci surgical system. A bespoke holder was fabricated to interface the robotic fiber with the da Vinci® instrument (Fig. \ref{fig: figure7}(a)). To simulate a laser delivery tool, we inserted a single-mode optical fiber through the robotic fiber and connected a red LED to its proximal end to simulate laser ablation.

During the phantom experiment, three da Vinci instruments were used concurrently: a laparoscopic camera for imaging, a grasper for manipulating the phantom tissue, and the fiberbot with da Vinci instrument assembly for distal scanning. The da Vinci system was used to introduce the robotic fiber into the endoluminal phantom and to position the fiber tip near the simulated tumor region (Fig. \ref{fig: figure7}(b)). The operator then used the grasper to peel back the phantom's surface layer, exposing the black target region (Fig. \ref{fig: figure7}C). While observing the laparoscopic video on the monitor, the operator controlled the robotic fiber using a joystick and guided the tip to scan along the tumor boundary (Fig. \ref{fig: figure7}(d)). The resulting motion path of the fiberbot tip is shown in Figure \ref{fig: figure7}(e). The complete scanning task required approximately 4 minutes.

\section{DISCUSSION}
In this work, we present a thermally actuated robotic fiber capable of partially actuating a segment along its length. This approach is particularly beneficial for MIS, as it provides a passive fiber segment that can pass through the working channel of a laparoscopic instrument with minimal Joule heating or coupled motion between the tip and the shaft. Additionally, the fiber enables precise motion, potentially enhancing precision in surgical procedures. 

This paper describes the robotic fiber's working principle, fabrication methods, electronic and handle assembly, and actuation system. 
We experimentally characterized the fiberbot's workspace, input–displacement relationship, path-following performance, and thermal response. Thermal camera imaging of the fiberbot further confirmed successful selective actuation. As a pilot study, we integrated the selectively actuated fiber with a da Vinci instrument and demonstrated it using an abdominal phantom. 

While the fiberbot allows for precise tip positioning, this work has certain limitations related to both the robotic system and the evaluation experiments. Future work will focus on employing polymers with a higher glass transition temperature to fabricate the robotic fiber, as this allows more power to be delivered to the fiber and thus increases motion range, while considering safety due to elevated temperatures. Additionally, we will replace the current Kapton tape protection at the fiber tip with a polymer enclosure to ensure reliable electrical insulation in liquid environments.



\section{CONCLUSION}
In this work, we presented a thermally actuated robotic fiber with a selective actuation mechanism for high-precision MIS applications. The fiber is fabricated using thermal drawing technology, enabling scalable and precise integration of metallic wires within a compact, flexible structure. We characterized and evaluated its motion behavior and thermal response through both simulations and experimental studies, demonstrating the system’s linearity and control accuracy. A proof-of-concept phantom study further confirmed the device’s effectiveness in navigating and performing scanning within an abdomen model. The proposed robotic fiber system has the potential to enhance the precision and efficiency of laparoscopic surgical tools.


\section*{APPENDIX}
 \begin{itemize}
 \item Supplementary video: An abdominal phantom study demonstrating its use in scanning along a simulated tumor.
 
 \end{itemize}

\section*{ACKNOWLEDGMENT}

The work of Jinshi Zhao was supported by the CRUK Grand Challenge Rosetta no. C59824/A25044, and the CRUK Convergence Science Centre at The Institute of Cancer Research, London, and
Imperial College London A26234. The work of Libaihe Tian was supported by the Engineering and Physical Science Research Council UK grant EP/W004224/1. The work of Alex Ranne is supported by the InnoHK initiative of the Innovation and Technology Commission of the Hong Kong Special Administrative Region Government.

\label{Section: Conclusions}

\bibliographystyle{unsrt}
\bibliography{refs}
\vfill
\newpage

\end{document}